# K-Means Based TinyML Anomaly Detection and Distributed Model Reuse via the Distributed Internet of Learning (DIoL)

Abdulrahman Albaiz
Department of Computer Science
& Engineering
Wright State University
Dayton, Ohio, USA
Email: albaiz.2@wright.edu

Fathi Amsaad
Department of Computer Science
& Engineering
Wright State University
Dayton, Ohio, USA
Email: fathi.amsaad@wright.edu

*Abstract*— **This paper presents a lightweight K-Means anomaly detection model and a distributed model-sharing workflow designed for resource-constrained microcontrollers (MCUs). Using real power measurements from a mini-fridge appliance, the system performs on-device feature extraction, clustering, and threshold estimation to identify abnormal appliance behavior. To avoid retraining models on every device, we introduce the Distributed Internet of Learning (DIoL), which enables a model trained on one MCU to be exported as a portable, text-based representation and reused directly on other devices. A two-device prototype demonstrates the feasibility of the "Train Once, Share Everywhere" (TOSE) approach using a real-world appliance case study, where Device A trains the model and Device B performs inference without retraining. Experimental results show consistent anomaly detection behavior, negligible parsing overhead, and identical inference runtimes between standalone and DIoL-based operation. The proposed framework enables scalable, low-cost TinyML deployment across fleets of embedded devices.**

*Keywords*— *TinyML, anomaly detection, microcontroller, K-Means clustering, embedded machine learning, DIoL, distributed learning.*

## I. Introduction

Power consumption patterns of electrical devices and appliances exhibit characteristic side-channel signatures that reflect their internal operating states. Monitoring these signatures can enable early detection of abnormal behavior—such as prolonged active states, irregular cycling, or unexpected shutdowns—which often precede mechanical faults, degraded efficiency, or unsafe operating conditions. For low-cost and distributed environments such as homes, pharmacies, and small warehouses, anomaly detection is often required to operate directly on microcontrollers (MCUs) to avoid reliance on cloud connectivity, continuous supervision, or external training resources [1]–[3]. Prior work has examined appliance behavior and power-related anomaly characteristics in real-world monitoring and smart-environment settings [4], [5]. In this work, a mini-fridge appliance is used as a representative real-world case study to demonstrate the feasibility of the proposed approach on a practical, resource-constrained system.

However, deploying data-driven anomaly detection on MCUs presents three fundamental challenges. First, training models on-device is computationally expensive and energy-intensive, particularly when learning from multi-day time-series data. Second, MCUs typically lack widely adopted standardized mechanisms for storing and reusing trained TinyML models across devices, forcing each device to retrain locally—even when identical appliances operate under similar conditions, as observed in prior MCU-based anomaly detection studies [6]–[8]. Third, existing federated and distributed learning approaches assume the presence of servers, wireless connectivity, or sufficient memory to maintain model updates, making them unsuitable for constrained standalone MCUs due to their communication and memory requirements [9], [10].

This paper addresses these limitations by introducing a lightweight clustering-based anomaly detection method and a practical mechanism for distributed model reuse across devices.

Our contributions are threefold:

*1) On-MCU K-Means Anomaly Detection:* We implement a resource-efficient K-Means clustering model that trains directly on the MCU using five power-based features (RMS, rolling mean, standard deviation, RMS slope, and compressor ON duration). The model is trained using approximately 20% of the collected dataset, selected chronologically from an initial period of stable, normal appliance operation to ensure representative baseline behavior while limiting on-device training cost. The trained centroids and normalization statistics are exported to a portable model file.

*2) Portable MODEL.TXT Format for Local and Cross-Device Reuse:* The MCU automatically generates a standardized MODEL.TXT file containing cluster centroids, feature scaling parameters, and decision thresholds. Any other device can load this file from local storage and perform inference without retraining. This enables "Train Once, Share Everywhere" operation on low-resource devices.

*3) Distributed Internet of Learning (DIoL) Prototype for Model Sharing:* We demonstrate the first DIoL-style workflow for TinyML on MCUs: a two-device experiment in which Device A performs training and Device B loads the generated

MODEL.TXT and directly runs inference. This prototype validates the feasibility of distributed model reuse and provides a foundation for future cloud-based model exchange and secure update mechanisms.

## II. RELATED WORK

### A. K-Means and Clustering-Based Anomaly Detection on Embedded Devices

Classical unsupervised anomaly detectors such as Isolation Forest and Local Outlier Factor (LOF) have been widely used in generic anomaly detection settings [11], [12], but their computational and memory demands make them less attractive for embedded MCUs. In contrast, K-Means is widely used in unsupervised anomaly detection due to its simplicity and effectiveness [13]. Prior embedded implementations have focused on reducing memory footprint, limiting iterations, or performing approximate updates to accommodate microcontroller constraints. However, existing work typically assumes offline training or requires external compute resources for model generation. In contrast, our approach performs both training and inference fully on the MCU and exports the learned centroids for reuse across devices.

### B. TinyML and Resource-Constrained Learning

Recent advancements in TinyML have enabled lightweight models to run on MCUs with only a few kilobytes of memory [1]–[3]. Prior research has explored on-device inference for tasks such as classification, regression, and anomaly detection, often relying on pre-trained models or external tooling for training [4], [6]. However, end-to-end on-device training remains uncommon due to memory and compute limitations on resource-constrained devices [7], [8]. Our work contributes to this gap by demonstrating a fully on-MCU training pipeline for multivariate clustering using real-world time-series data.

### C. Distributed and Federated Learning

Federated learning allows models to be trained collaboratively across multiple devices while keeping raw data local, but these techniques assume the availability of communication infrastructure, sufficient memory for model updates, and iterative global aggregation—requirements incompatible with low-power MCUs for TinyML anomaly detection [9], [10]. Lightweight variants exist but still depend on periodic connectivity and server coordination. Our approach differs fundamentally: rather than distributing training, we enable model reuse by allowing devices to exchange a compact portable model representation without synchronization rounds or cloud orchestration.

### D. Internet of Learning (IoL) and Knowledge Sharing Across Devices

The emerging IoL concept envisions decentralized sharing of learned models or parameters among devices. Existing work largely targets cloud-capable or mobile platforms and has not addressed MCU-focused constraints such as microSD-based storage, sub-100 KB memory limits, or offline operation. Prior work on distributed or online model updates has primarily targeted cloud-capable systems rather than resource-constrained MCUs [9], [10]. Our work introduces a practical realization of IoL for TinyML through the Distributed Internet of Learning (DIoL), enabling offline sharing and reuse of anomaly detection models across microcontroller-based devices.

### E. Gap in the Literature

To the best of our knowledge, no prior work has demonstrated:

(1) a complete on-MCU K-Means training pipeline for multivariate appliance-side power features;

(2) a standardized portable model format enabling training-skipping inference on other MCUs; and

(3) an operational DIoL prototype demonstrating cross-device model reuse.

This combination positions our work as the first practical framework for distributed TinyML anomaly detection on resource-constrained embedded systems.

## III. SYSTEM OVERVIEW

The proposed system performs end-to-end power-side-channel anomaly detection using a microcontroller-based platform. The hardware setup consists of an STM32F446RE MCU, an ACS712 Hall-effect current sensor for real-time current measurement, and a microSD card for local data storage [14], [15]. This configuration enables continuous sensing, feature extraction, model training, and inference without external computational resources or network connectivity.

The MCU samples the current waveform, computes RMS values over fixed windows, and logs timestamped measurements to the microSD card. A multi-stage processing pipeline then runs entirely on the device over logged data: data cleansing, feature extraction, state labeling, model training, and anomaly detection. Five features derived from the time-series data—RMS, rolling mean, standard deviation, RMS slope, and compressor ON duration—serve as input to clustering.

To support distributed model reuse, we adopt the Distributed Internet of Learning (DIoL) paradigm, introduced in this work, which enables an MCU to train a model once and share it with other devices through a portable model file stored on microSD. This approach follows the "Train Once, Share Everywhere" (TOSE) principle, allowing a model trained on one MCU to be directly reused on others without additional training. In this paper, DIoL is demonstrated using a two-device workflow: Device A trains a K-Means model and exports it as a portable file, while Device B loads the file and performs inference.

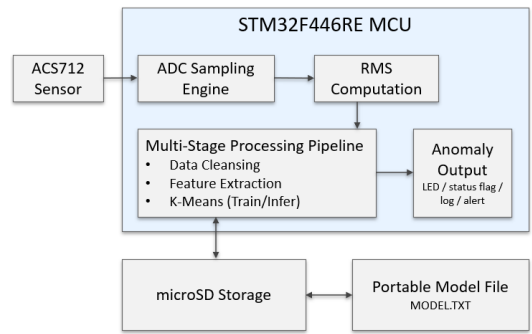

Fig. 1. System architecture and DIoL workflow overview.

Together, the hardware platform, real-world data pipeline, and DIoL workflow provide a practical foundation for scalable, low-cost anomaly detection across fleets of embedded devices. The integrated hardware and processing pipeline is illustrated in Fig. 1.

## IV. K-Means Model Design for MCU

K-Means clustering provides a simple unsupervised alternative to more complex outlier detection models while remaining suitable for resource-constrained embedded platforms [13]. To formalize, K-Means partitions feature vectors into $k = 3$ clusters by minimizing the within-cluster sum of squared distances:

$$arg \min_{\{\mu_i\}_{i=1}^k} \sum_{i=1}^{k} \sum_{x \in C_i} \| x - \mu_i \|^2$$

where $\mu_i$ denotes the centroid of cluster $C_i$. Each feature vector is assigned to the nearest centroid based on Euclidean distance.

The choice of $k = 3$ reflects the dominant operating regimes observed in the appliance power signal, corresponding to compressor off, steady operation, and transient or elevated-load behavior. This selection balances representational fidelity and model simplicity, avoiding over-segmentation while preserving stable centroid estimation under limited training data. Informal evaluation with alternative $k$ values did not yield improved separation of normal operating behavior for this appliance, reinforcing the suitability of $k = 3$ under the given constraints.

The K-Means model is implemented as a lightweight, fully on-MCU algorithm tailored to the memory and compute constraints of the STM32F446RE. The model operates on five features extracted from the power measurement stream: RMS current, rolling mean, standard deviation, RMS slope, and compressor ON duration. These features capture both instantaneous behavior and longer-term temporal patterns of appliance operation.

To minimize computational cost and energy consumption, the MCU trains the model using approximately 20% of the collected dataset, selected chronologically from an initial period of stable, normal appliance operation. This subset provides a representative baseline while limiting on-device training overhead. Each feature is normalized using mean and standard deviation values computed during this initial pass. Centroids are initialized from the first $k$ scaled samples, and a small, fixed number of Lloyd's iterations is performed to update centroids, ensuring deterministic memory usage and avoiding dynamic allocation.

For inference, each incoming feature vector is normalized using the stored statistics and assigned to the nearest centroid. An anomaly score is computed as the Euclidean distance to that centroid. The anomaly threshold is derived from the 95th percentile of training distances computed over the baseline subset, providing a conservative boundary for normal operation under constrained training data. A tunable scaling factor is applied to adjust sensitivity based on deployment requirements.

Upon completing training, the MCU serializes the learned centroids, normalization statistics, and anomaly threshold into a portable model file. This file enables direct reuse of the trained model by other devices without retraining, forming the basis of the Distributed Internet of Learning (DIoL) workflow described in the following section. The MCU implementation of the K-Means training and inference pipeline is illustrated in Fig. 2.

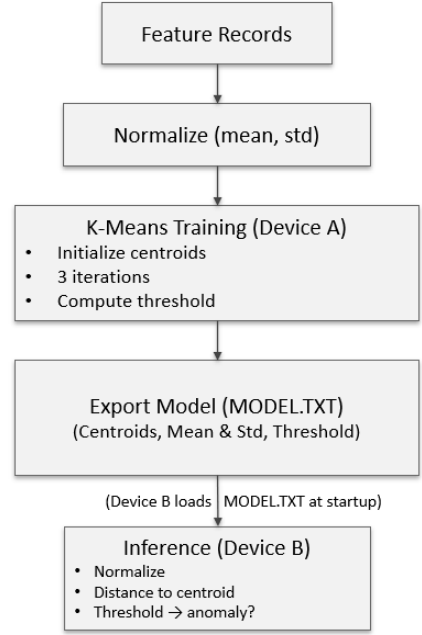

Fig. 2. K-Means training and inference pipeline.

## V. Portable Model File Structure and Sharing

After completing K-Means training, the MCU serializes the learned model into a compact, portable model file stored on the microSD card. This file contains all information required for standalone inference on any compatible device, including the number of clusters, centroid coordinates, feature normalization statistics, and the anomaly detection threshold. The representation is lightweight, human-readable, and designed to be parsed efficiently on memory-constrained microcontrollers.

The file is organized into labeled sections to simplify loading and validation. Centroids are stored row-wise, with each line representing one cluster and containing the normalized feature values learned during training. The mean and standard deviation vectors are included to ensure consistent feature scaling across devices, enabling a model trained on one MCU to be applied directly on another without recalibration. The threshold section encodes the empirically derived distance boundary used to identify anomalous points during inference.

To reuse the model, a second MCU simply reads the portable file from its microSD card and loads the parameters into memory at startup. Inference then proceeds without performing any local training, eliminating computational overhead and enabling rapid deployment across multiple appliances. This design forms the core of the DIoL workflow, enabling practical cross-device model reuse on low-power embedded systems without cloud connectivity or coordinated updates.

## VI. DIoL Workflow Demonstration (Prototype)

We introduce the Distributed Internet of Learning (DIoL) as a paradigm for sharing TinyML models across MCUs. To demonstrate feasibility, a two-device prototype is implemented consisting of a training device (Device A) and an inference-only device (Device B). Both devices are based on the STM32F446RE platform and use identical sensing and storage configurations. Device A trains the K-Means model and exports a portable model file, while Device B loads this file and performs inference without any local training. The DIoL workflow is summarized in Fig. 3.

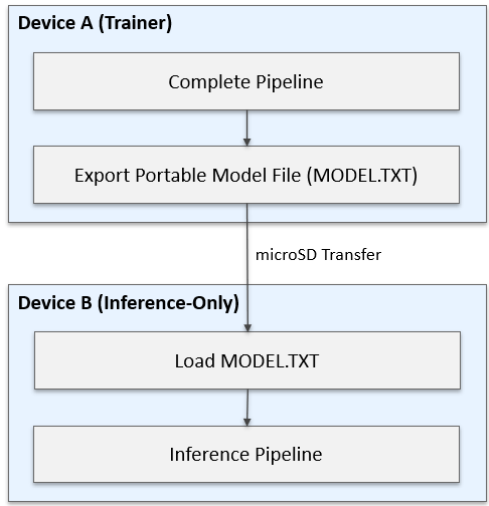

Fig. 3. DIoL workflow: Train Once, Share Everywhere.

### A. Device A — Model Training and Export

Device A executes the full pipeline, including feature extraction, K-Means training, and threshold estimation. Upon completion, it generates a portable model file containing the centroids, feature scaling parameters, and anomaly threshold, which is stored on the microSD card for transfer.

### B. Device B — Training-Skipping Inference

Device B loads the model file at startup and initializes its anomaly detection pipeline without performing local training. Incoming feature vectors are normalized using stored statistics and classified based on distance to the pre-trained centroids, demonstrating immediate operation with reduced setup time and energy consumption.

### C. End-to-End Operation and Results

The DIoL prototype reproduces the detection behavior of the training device, confirming that portable model files preserve consistency across devices. Runtime measurements show no additional inference overhead beyond the initial model loading step. During model loading, the system performs basic validation checks to verify parameter counts and value ranges. In the event of corrupted or incomplete files, the model is safely rejected and the device continues operating with its existing configuration. More advanced integrity protection mechanisms are reserved for future work.

## VII. Experimental Evaluation

We evaluated the proposed K-Means anomaly detection model and DIoL workflow using fourteen days of real power measurements collected from a mini-fridge appliance as a focused case study for on-device TinyML deployment. The MCU processed 43,285 feature records, with approximately 8,600 samples used for training and the remainder for inference. Our evaluation assessed detection performance, runtime efficiency, resource usage, and the effectiveness of training-skipping reuse through the DIoL workflow.

### A. Detection Performance

The K-Means model successfully detected all injected abnormal events in the evaluated dataset, including extended compressor runtimes, short cycles, and prolonged power-off periods. Detected anomalies aligned with the injected ground truth intervals, indicating robust discrimination between normal and abnormal behavior under natural compressor variability. Based on the injected anomaly labels, the detector achieved full recall for the evaluated events while maintaining a low incidence of false positives during normal operation. A timeline illustrating typical normal compressor cycles and duration-based anomalies is shown in Fig. 4.

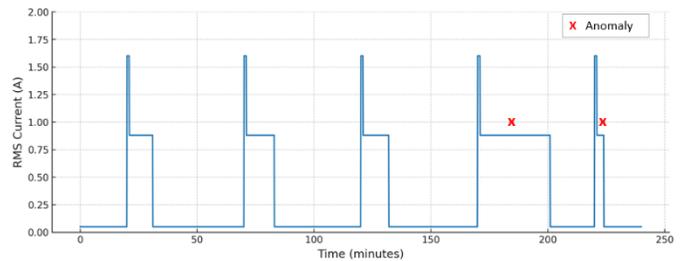

Fig. 4. Compressor behavior timeline with anomalies marked.

### B. Comparison with Z-Score

Both K-Means and a previously developed Z-Score–based detector exhibited similar detection behavior, identifying the same anomalous intervals; however, K-Means produced smoother decision boundaries due to its clustering structure and multivariate distance metric. Runtime performance for Z-Score inference was 58,953 ms for 43,285 records (734 records/s), while K-Means achieved comparable throughput with an inference runtime of 58,518 ms. K-Means training required 59,328 ms and was performed once per device or skipped entirely through DIoL-based reuse.

In addition to runtime performance, both models exhibit comparable memory footprints, relying on fixed-size statistics and avoiding dynamic allocation. Energy consumption during inference is similarly dominated by lightweight feature extraction and distance computation. As a result, neither model introduces additional overhead that would limit deployment on resource-constrained MCUs.

## C. Model File Size and Parsing Overhead

The portable model file generated after training is compact, storing only cluster centroids, normalization parameters, and a threshold (e.g., threshold = 8.99679 for K=3 clusters). Parsing the file on a second MCU required negligible overhead, enabling inference to begin immediately at startup.

## D. DIoL Training-Skipping Evaluation

Using the DIoL workflow, Device B loaded Device A's model and performed inference without retraining. The DIoL inference runtime was 58,514 ms, essentially identical to standalone K-Means inference, confirming that model reuse does not introduce additional computation costs. The detected anomalies from DIoL inference matched those from Device A exactly, demonstrating consistent behavior and validating cross-device portability.

## E. Memory and Runtime Constraints

The implementation operated within the STM32F446RE's resource limits, using fixed-size arrays, three K-Means iterations, and a 20% training data subset to ensure deterministic memory usage. Both models fit comfortably within flash and SRAM budgets, satisfying deployment requirements for low-power embedded environments.

## VIII. DISCUSSION

The results demonstrate that distributed model reuse through DIoL provides a practical and scalable mechanism for deploying TinyML anomaly detection across fleets of embedded devices. By training a single device and distributing a portable model file, redundant computation is eliminated while maintaining consistent detection behavior across identical or functionally similar devices deployed in household or storage environments.

A key strength of the proposed approach is its generality. Although evaluated using K-Means clustering—and previously with a Z-Score–based detector—the DIoL workflow is model-agnostic and can support other lightweight techniques that can be serialized and loaded within MCU memory constraints. Separating training from inference allows learning to be performed once, avoiding repeated energy and compute costs on deployed devices.

Training on a limited data subset reflects realistic autonomous MCU constraints and was sufficient for stable centroid and threshold estimation in this appliance. While broader datasets and alternative sampling strategies may improve adaptability in more dynamic settings, this work prioritizes deterministic execution and practical feasibility. Evaluation was conducted on a single appliance over fourteen days to emphasize reproducibility under realistic constraints; extension to additional devices, environments, and secure model-sharing mechanisms is left for future work.

## IX. CONCLUSION & FUTURE WORK

This paper presented a lightweight K-Means–based anomaly detection system and a Distributed Internet of Learning (DIoL) workflow enabling cross-device model reuse on resource-constrained microcontrollers. Using real power side-channel measurements from a mini-fridge appliance, the system performed on-device feature extraction, training, inference, and model serialization on an STM32 MCU. A second device successfully executed training-skipping inference using the exported model, achieving identical detection performance with no additional computational overhead.

Future work will extend DIoL beyond local microSD-based transfer toward network-assisted and secure model sharing to support large-scale device fleets. Additional evaluation across diverse appliance types and operating conditions will further assess robustness, scalability, and long-term stability.

ACKNOWLEDGMENT

This work was supported by the Saudi Arabian Cultural Mission (SACM).

REFERENCES

[1] Y. Abadade, A. Temouden, H. Bamoumen, N. Benamar, Y. Chtouki and A. S. Hafid, "A Comprehensive Survey on TinyML," IEEE Access, vol. 11, pp. 96892–96922, 2023.

[2] R. Sanchez-Iborra and A. F. Skarmeta, "TinyML-Enabled Frugal Smart Objects: Challenges and Opportunities," IEEE Circuits and Systems Magazine, vol. 20, no. 3, pp. 4–18, 2020.

[3] C. R. Banbury, V. J. Reddi, M. Lam, W. Fu, A. Fazel, J. Holleman, X. Huang, R. Hurtado, D. Kanter, A. Lokhmotov, D. Patterson, D. Pau, J.-S. Seo, J. Sieracki, U. Thakker, M. Verhelst and P. Yadav, "Benchmarking TinyML Systems: Challenges and Direction," arXiv:2003.04821, 2020.

[4] D. Fährmann, L. Martín, L. Sánchez and N. Damer, "Anomaly Detection in Smart Environments: A Comprehensive Survey," IEEE Access, vol. 12, pp. 64006–64049, 2024.

[5] M. Zeifman and K. Roth, "Nonintrusive Appliance Load Monitoring: Review and Outlook," IEEE Transactions on Consumer Electronics, vol. 57, no. 1, pp. 76–84, 2011.

[6] K. Hasegawa, K. Chikamatsu and N. Togawa, "Empirical Evaluation on Anomaly Behavior Detection for Low-Cost Micro-Controllers Utilizing Accurate Power Analysis," Proc. IEEE Int. Symp. On-Line Testing and Robust System Design (IOLTS), pp. 54–57, 2019.

[7] M. Antonini, M. Pincheira, M. Vecchio and F. Antonelli, "A TinyML Approach to Non-Repudiable Anomaly Detection in Extreme Industrial Environments," Proc. IEEE Int. Workshop on Metrology for Industry 4.0 & IoT (MetroInd4.0&IoT), pp. 397–402, 2022.

[8] M. Lord and A. Kaplan, "Mechanical Anomaly Detection on an Embedded Microcontroller," Proc. 2021 Int. Conf. Computational Science and Computational Intelligence (CSCI), pp. 562–568, 2021.

[9] D. Pau, A. Khiari and D. Denaro, "Online Learning on Tiny Micro-Controllers for Anomaly Detection in Water Distribution Systems," Proc. IEEE Int. Conf. Consumer Electronics (ICCE-Berlin), pp. 1–6, 2021.

[10] M. Cardoni, D. P. Pau, L. Falaschetti, C. Turchetti and M. Lattuada, "Online Learning of Oil Leak Anomalies in Wind Turbines with Block-Based Binary Reservoir," Electronics, vol. 10, no. 22, p. 2836, 2021.

[11] F. T. Liu, K. M. Ting and Z.-H. Zhou, "Isolation Forest," Proc. IEEE Int. Conf. Data Mining (ICDM), pp. 413–422, 2008.

[12] M. M. Breunig, H.-P. Kriegel, R. T. Ng and J. S. Sander, "LOF: Identifying Density-Based Local Outliers," Proc. ACM SIGMOD Int. Conf. Management of Data, pp. 93–104, 2000.

[13] V. Chandola, A. Banerjee, and V. Kumar, "Anomaly Detection: A Survey," ACM Computing Surveys, vol. 41, no. 3, pp. 1–58, 2009.

[14] STMicroelectronics, "STM32F446xC/E Datasheet," Rev. 10, Jan. 2021. [Online]. Available: https://www.st.com/resource/en/datasheet/stm32f446re.pdf.

[15] Allegro MicroSystems, "ACS712: Fully Integrated, Hall-Effect-Based Linear Current Sensor IC," Datasheet. [Online]. Available: https://www.allegromicro.com/-/media/files/datasheets/acs712-datasheet.ashx.